# Towards an Automatic Recognition of Mixed Languages in R: The Case of Ukrainian-Russian Hybrid Language Surzhyk

**Nataliya Sira[1], Giorgio Maria Di Nunzio[2,3], Viviana Nosilia[1]**

[1]Department of Linguistic and Literary Study, [2]Department of Information Engineering, [3]Department of Mathematics
University of Padua, Italy
nsira28@gmail.com, giorgiomaria.dinunzio@unipd.it, viviana.nosilia.unipd.it

## Abstract

Language interference is common in today's multilingual societies where more languages are being in contact and as a global final result leads to the creation of hybrid languages. These, together with doubts on their right to be officially recognised, made emerge in the area of computational linguistics the problem of their automatic identification and further elaboration. In this paper, we propose a first attempt to identify the elements of a Ukrainian-Russian hybrid language, Surzhyk, through the adoption of the example-based rules created with the instruments of programming language R. Our example-based study consists of: 1) analysis of spoken samples of Surzhyk registered by Del Gaudio (2010) in Kyiv area and creation of the written corpus; 2) production of specific rules on the identification of Surzhyk patterns and their implementation; 3) testing the code and analysing the effectiveness of the hybrid language classifier.

## 1. Introduction

The phenomena of code mixing might be considered a natural product of multilingualism. According to Barman, Das, Wagner, Foster (2014: 13) in the situations where speakers switch between languages or mix them, the automatic language identification process is increasingly important as it facilitates further language processing. Being Surzhyk a hybrid language that involves Ukrainian and Russian languages in its creation, the automatic language identification, which would allow then Surzhyk processing, requires great efforts.

The hybridity of the mixed languages consists in the fact that they take their lexicon from one source and grammar from another. Their classification becomes difficult as on the basis of the lexicon they could appertain to one language family and on the basis of morphology, syntax and general grammatical characteristics they may belong to another language family (Bakker, Mous, 1994: 5). According to the criterion aimed to define mixed language nature provided by Bakker in 1992, "a language is mixed if it can with equal justification be assigned to two different language families." (Bakker 1992 in Bakker, 1994: 26).

In the current research the idea of the Matrix Language Frame model (MLF) elaborated by Myers-Scotton (2002, 2003) and proposed by Kent (2010) for Surzhyk analysis is taken into consideration. According to the MLF model, only one language is the source of the abstract morphosyntactic frame in a bilingual clause and that is the Matrix Language (ML). The other participating language is the Embedded Language (EL) and it must agree with structural requirements stipulated by the ML (Myers-Scotton 2002, 2003). The nature of Surzhyk is complex: in Ukrainian-Russian Surzhyk structure, Ukrainian is a ML whereas Russian is an EL. In addition, it has a huge variety of possible mutation in base of the geographical area and speaker's characteristics. This spoken language is commonly used in the whole Ukraine. The territories of the country where was registered the largest number of Surzhyk speakers were eastern, southern and central parts of Ukraine. (Vakhtin et al., 2003 in Kent, 2010: 43).

### 1.1 Objective of the Study

Since the development of mixed languages is rapid and irreversible, the study aims to demonstrate first attempts on the automatic identification of such languages. In our case, we propose an example-based study of Surzhyk samples recorded in different areas of Ukraine and their elaboration in the R programming language aimed to identify particular patterns of Surzhyk verbs. After the creation of the text corpus of Surzhyk, we studied the terminology present in this corpus and produced the terminological tables that we used to identify the patterns of the Surzhyk language. After a careful selection of the patterns, we implemented the classification rules in R and we tested the efficiency of this classification. The corpus as well as the source code is available on Github for reproducibility purposes.

The remainder of the paper is organized as follows. In Section 2, we present the preliminary analysis on the definition of patterns in Surzhyk; in Section 3, …

## 2. Preliminary Analysis

In this section, we present background studies on Surzhyk patterns definition. The decision to analyse two particular characteristics of Surzhyk verbs was taken with the regard on the recurring repetition of these in the processed texts that was noticed during the process of manual transcription of the registrations collected by Del Gaudio (Del Gaudio, 2010). Firstly, we analysed the ending "-м" in the first person plural of the Present tense and secondly, the prefix "под-" of Surzhyk verbs. To carry out the analysis of the first particle, it was necessary to compare the Russian and Ukrainian verbal systems, and, specifically, the ending of the first person plural in the Present tense, together with its historic development. Regarding the analysis of the second particle of interest, we took into consideration the word formation system in Russian and in Ukrainian and compared prefixes that are usually adopted in prefixation process.

## 2.1 The Study of the First Person Plural Ending Form "-м"

In Ukrainian and in Russian, there are two verb conjugations: Conjugation 1 and Conjugation 2.[1] The following table (Table 1) compares the Present tense endings in Russian and Ukrainian languages.[2]

| Russian | | Ukrainian | |
|---|---|---|---|
| Conj. 1 | Conj. 2 | Conj. 1 | Conj. 2 |
| -ю (-у) | -ю (-у) | -у (-ю) | -у (-ю) |
| -ешь (ёшь) | -ишь | -еш (-єш) | -иш (-їш) |
| -ет (-ёт) | -ит | -е (-є) | -ить (-їть) |
| -ем (-ём) | -им | -емо (-ємо) | -имо (-їмо) |
| -ете (-ёте) | -ите | -ете (-єте) | -ите (-їте) |
| -ют (-ут) | -ят (-ат) | -уть (-ють) | -ать (-ять) |

Table 1: Present tense endings in Russian and Ukrainian

In the first person plural of the Present tense, a vowel that we have considered as a part of the ending in the table, from another point of view can be seen as a thematic or connector vowel between a stem and an ending of a verb. Consequently, we can admit that possible endings of the Present tense in first person plural can be **-мо** in standard Ukrainian or **-м** in standard Russian. The difference that nowadays seems to be clear in two standards was not so definite before, especially for the Ukrainian language. A Ukrainian grammar of 1927 presents the ending **-м** as possible variant to the ending **-мо** in the first person plural. (Buzuk 1927 in Horbach 1985, 81). For space limits we do not present a description of historical development of these endings that may be consulted in Filin (1972, 463) and in Kuznetsov (1953, 206).

## 2.2 The Study of the Prefix "под-"

While in Russian language there is a prefix **под-** that partially corresponds in the meaning to the preposition **под** (which means "under") (Grammatika Russkogo Jazyka 596-597), modern Ukrainian language does not present prefix "под-". A regular Ukrainian prefix corresponding to the prefix **под-** we analysed in the Surzhyk samples would be a prefix **під-** that is one of the most important prefixes involved in verbal word formation process. (Ukrajinska mova, 347-348). It should be underlined that in the Ukrainian lexicon there are words with a **по-** prefix and, when followed by a consonant **д**, at first glance they may deceive a non specialist of the area, especially when the morphology of the word is not taken into consideration or neglected. The prefix "под-" results to be Russian but in combination to Ukrainian verbs seems to be a significant and characteristic particle in the identification of Surzhyk lexicon.

## 3. Data Collection

Since Surzhyk is more characteristic as a spoken language rather than written, the collection of Surzhyk samples was a critical task. During the first stage of the research we have analysed the interviews recorded by Del Gaudio in different areas of Ukraine involving people from Kyiv, Chernihiv and Kharkiv areas. In the current research, only the first group of conversations regarding the area of Kyiv were elaborated in detail. In order to have the digital version of these conversation, it was necessary to transcribe manually all the registrations since no effective software for speech-to-text automatic recognition was available neither for Ukrainian nor for a non-standard spoken language Surzhyk. The first stage of the analysis consisted thus in creating the Surzhyk corpus. Secondly, in the previously created text documents we identified lexical elements that did not belong to the standard Ukrainian language. Finally, we created and filled the terminological records that in total present more than 1,000 non-standard Ukrainian terms. Although creating terminological tables for Surzhyk was only part of the preparatory phase and not the main task of this research, we consider that they may be useful for further studies on Surzhyk.

## 4. Rules for the Automatic Identification of Surzhyk Patterns

In this section, we present the linguistic patterns that we identified in order to automatically identify the Surzhyk hybrid language.

## 4.1 Final pattern "-м" characterising the first person plural verbs in Present tense

Two groups of rules were created: general and specific. Subsequently, we present the rules of both groups in the summarising form. In Table 2 and Table 3, we show an example of a set of "general" rules and "specific" rules on the identification of patterns in the Surzhyk language.

---

[1] This part, including the table, is based on: Akademija Nauk SSSR 1980: 647, 663 and Ukrajinskyj pravopys 2015: 108-117 that can be consulted for more details.

[2] In this paper "Present" is used conventionally, due to the space limits, but it always intends to refer to "Present / Future", depending on the verb aspect.

In Table 2, the first two rules determine that the first element is a word "ми" or "самі" while the second element "-м" is a final part of the word. The second element follows the first and is situated on max distance 3 from the first element. The second two rules imply that the first element ends in "-м" while the second one ends in "-ти" or "-ть" and is situated on max distance 3 from the first element. The general rules present also requirements which has to be respected by the output word. In the case of the first general rule the first pattern of the output has to be a single word "ми" and not a part of a word; the similar requirement regards also the second pattern of the third rule: the second pattern of the output that ends in "-ти" has to be a word with a number of characters higher than the number of characters present in the word "ти" since it would be a personal pronoun of the second person singular and not a verb ending.

Similarly to the general rules, these specific rules, shown in Table 3, also have requirements to be fulfilled by the output word. For example, the rule n. 4 requires the second element of the output that ends in "-їм" to be a word with a number of characters higher than the number of characters present in the word "їм". This type of requirement is also present in the rules n. 8, 12, 16 and was necessary to prevent the false positive outputs that are personal pronouns of the third person plural in Dative case "їм". A similar requirement is present in the rules n. 9, 10, 11, 12: the output containing a final particle "-ти" has to be a word with a number of characters higher than the number of characters present in the word "ти". This excludes the false positive output "ти" as a single word expressing the personal pronoun of the second person singular.

| Rule num. | First pattern | Second pattern | Distance | Output word requests |
|---|---|---|---|---|
| 1 | ми | -м | max. 3 | 1st patt. = "ми" |
| 2 | самі | -м | max. 3 | |
| 3 | -м | -ти | max. 3 | 2nd patt. > "ти" |
| 4 | -м | -ть | max. 3 | |

Table 2: General rules on identification

| Rule num. | First pattern | Second pattern | Distance | Output word requests |
|---|---|---|---|---|
| 1 | ми | -ем | max. 3 | |
| 2 | ми | -єм | max. 3 | |
| 3 | ми | -им | max. 3 | |
| 4 | ми | -їм | max. 3 | 2nd patt. > "їм" |
| 5 | самі | -ем | max. 3 | |
| 6 | самі | -єм | max. 3 | |
| 7 | самі | -им | max. 3 | |
| 8 | самі | -їм | max. 3 | 2nd patt. > "їм" |
| 9 | -ем | -ти | max. 3 | 2nd patt. > "ти" |
| 10 | -єм | -ти | max. 3 | 2nd patt. > "ти" |
| 11 | -им | -ти | max. 3 | 2nd patt. > "ти" |
| 12 | -їм | -ти | max. 3 | 1st. patt. > "їм" <br> 2nd patt. > "ти" |
| 13 | -ем | -ть | max. 3 | |
| 14 | -єм | -ть | max. 3 | |
| 15 | -им | -ть | max. 3 | |
| 16 | -їм | -ть | max. 3 | 1st. patt. > "їм" |

Table 3: Specific rules on identification

### 4.2 Initial pattern "под-" characterising the verb formation process

In order to define verbs with a prefix "под-", the pattern "под" has to be an initial part of the word. Additionally, the output word has to present a number of characters superior of 3, meaning it has to be a part of a verb and not a single preposition "под" that is composed of 3 characters. We may also define the personal pronouns that precede the verb but in our samples their presence was limited, in most cases personal pronouns were omitted or were expressed by a noun. Therefore we decided to have general rules on the identification of verbs containing the prefix "под-". Based on this general rule we can develop further restrictions in the future, once we would have more digital material to analyse.

### 4.3 Implementation in R

The code written in R allows to display if there are matches between the created rules and Surzhyk samples. An example of the creation of a rule in R can be presented as follows:

```
first_suffix <- #insert Surzhyk suffix #1
second_suffix <- #insert Surzhyk suffix #2
distance <-3
first_suffix_found <- tidy_interviews %>%
  filter(grepl(x = word,
               pattern = paste0(first_suffix, "$")))
```

```
print(first_suffix_found)
second_suffix_found <- tidy_interviews %>%
  filter(grepl(x = word,
                               pattern =
paste0(second_suffix, "$")))
print(second_suffix_found)
first_suffix_found %>%
  inner_join(y = second_suffix_found, by =
c("file", "line")) %>%
    rename(pos_first_suf  =  position.x,
first_word = word.x, # rename columns
             pos_second_suf = position.y,
second_word = word.y) %>%
  filter(pos_second_suf - pos_first_suf <=
distance  &  pos_second_suf  -  pos_first_suf
> 0) %>%
  inner_join(y = interviews)
```

This is an original structure we created and implemented in R. It can be used for general and specific rules by defining the first and second pattern "suffix". Notice that we adopted term "suffix" at the early stage of the code development and when searched in R it may be affix, pattern, final particle and not necessarily suffix. In some cases in place of "suffix" we may have "word".

The R source code and the full documentation of the specific rules is available online for reproducibility purposes.[3]

## 5. Analysis of the Results

In this section, we present the results for every combination of rules. During the testing phase, the rules were firstly tested on the previously analysed Surzhyk texts; secondly, the application on Surzhyk pattern identification was tested on the new texts. For this second part of testing we decided to select Russian texts of spoken language, in particular some transcribing of the real interviews present on the website of the *Radio svoboda* (*Radio Liberty*) in order to prove whether our rules will give some outputs. The outputs obtained during the testing process could be classified as "true positive" or "false positive". A "true positive" result is considered a term that corresponds to the aimed output criteria. For the identification of the verbs in the first person plural in Present tense ending in "-м" this means that the output entity has to be a Surzhyk verb in the first person plural in Present tense. When we refer to the identification of the verbs containing a prefix "под-", the output has to be a Surzhyk verb with a prefix "под-" and not a noun, for example.

### 5.1 General Rules Identification of the First Person Plural Verb in Present tense

The process of creating and testing the general rules may be considered as a necessary part of this research that led us to the creation of the specific rules. The output we received during the testing of these general rules on Surzhyk texts were mostly false positive and not related to the question of Surzhyk. Considering the low utility of the general rules (if compared to the specific rules), in this paper we decided to present the numbers of the outputs very briefly:
1) **ми + -м**: 10 results, 7 true positive;
2) **самі + -м**: 2 results, 1 true positive;
3) **-м + -ти**: 5 results, 5 false positive;
4) **-м + -ть**: 39 results, 3 true positive.

### 5.2 Specific rules outputs on the identification of the first person plural verb in Present tense

Since the specific rules regarding the identification of the first person plural verb in Present tense are more detailed, they allowed us to have more precise outputs. We identified 11 combinations of verbs with the first person plural ending. Among a total amount of 12 outputs only one result was a false positive. Consequently, we decided to present the outputs:

1) **ми + -ем**: 1 result, 1 true positive
(ми на нього кажем);
2) **ми + -єм**: 4 results, 4 true positive
(ми тут працюєм, ми взнаєм, ми чисто не балакаєм, хіба ми чисто балакаєм);
3) **ми + -им**: 1 result, 1 true positive
(ми лучшего на бачим);
4) **ми + -їм**: 1 result,1 true positive
(ми тебе устроїм);
5) **самі + -ем**: no results;
6) **самі + -єм**: 1 result, 1 true positive
(самі сієм);
7) **самі + -им**: no results;
8) **самі + -їм**: no results;
9) **-ем + -ти**: no results;
10) **-єм + -ти**: no results;
11) **-им +-ти**: no results;
12) **-їм + -ти**: no results;
13) **-ем + -ть**: 2 results, 2 true positive
(будем злазить, будем возвращать);
14) **-єм + -ть**: 1 result, 1 false positive
(вообщєм уже, да, двадцять);
15) **-им +-ть**: 1 result, 1 true positive
(мусим ходить);
16) **-їм + -ть**: no results.

### 5.3 General rule outputs on the identification of the verbs with the prefix "под-"

Testing the rule on the identification of the verbs containing a prefix "под-" gave us 8 outputs, of which 4 might be considered true positive; this means the output elements are verbs and "под-" is a prefix. As assumed, most of the false positive outputs are verbs with a prefix "по-" or, in the single case, a noun. The true positive outputs are verbs: "подожди", "подимаюся", "подработать" and "подвів".

---

[3] https://github.com/gmdn/Surzhyk

## 5.4 Analysis of additional testing of Surzhyk rules on Russian corpus

Surzhyk is a hybrid language between Russian and Ukrainian, which means that its element may also come from Russian. Consequently, we present short outline of the testing on Russian corpus.

### 5.4.1 General rules outputs on identification of the first person plural verb in Present tense

During the testing process of the general rules on Russian corpus we obtained 25 results; 20 of them were actually the outputs for the input search "-м" + "-ть" and 5 the results for the input "-м" + "ти". We did not analyse all the outputs in detail as it was concluded previously that the general rules were not effective for the identification of Surzhyk verbs of the first person plural. Secondly, it was decided to check if the specific rules may provide similar outputs. What was discovered by testing of the specific rules was that the total number of matches was 10. As supposed, the highest number of matches (8 of 10) corresponded to the input combination containing the pattern "-ем", which is common final ending for Russian, Ukrainian and Surzhyk words.

### 5.4.2 Specific rules outputs on the identification of the first person plural verb in Present tense

Three of the specific rules on identification of Surzhyk verbs led to the identification of the elements in the standard Russian texts. Some of these elements were verbs with the infinitive ending "-ть", others were irrelevant results because of their affiliation to the categories of Russian nouns, adjectives or personal pronouns that in different cases, such as Dative, Instrumental and Locative, present the final pattern "-ем" or "-им".

### 5.4.3 General rule outputs on the identification of the verbs with the prefix "под-"

When we tested the rule on the identification of the verbs with the prefix "под-" on Russian corpus, we obtained 28 outputs. Among these results are present verbs with a prefix "под-", verbs with a prefix "по-", nouns with a prefix "под-" or nouns that have "под" as an initial part of the stem.

## 6. Evaluation

Based on the results of the testing process we can proceed with the discussion on the efficiency of our rules. Firstly, we discuss the results of the study on the identification of Surzhyk verbs of the first person plural in Present tense. The specific rules allowed to identify 11 combinations of verbs with the first person plural ending. In a total amount of 12 outputs only one result was false positive. 11 results obtained with the general rules corresponded to the results we had with the specific rules. The difference consisted in the number of false positive outputs that were limited to 1 by adoption of the specific rules. It can be concluded that the rules we provided were effective if applied to our texts. Even though general rules can be seen only as a generalization of the specific rules and their outputs were mostly irrelevant in relation to our input and purpose of identifying Surzhyk verbs of the first person plural, they allowed to find more Surzhyk words and thus means they may be used to enlarge the corpus of Surzhyk.

In regard to the second group of elements of our interest, the verbs containing prefix "под-", we can see that the rule has to be improved in order to give a better performance. However, testing this rule on Surzhyk corpus gave us a couple of interesting verbs that do not appertain neither to Ukrainian nor to Russian vocabulary, among the results were also present verbs with a prefix "по-" followed then by a letter "д" as an initial consonant of a stem or of another prefix, or nouns.

As final phase of project development, it was decided to verify whether the rules on identification behave as expected on new texts. For this test, Russian texts of spoken language were selected. On the one side, one may suppose that no element should be identified as the rules we created are aimed to identify Surzhyk elements. On the other side, one may consider possible to have patterns we identified as Surzhyk within Russian text, especially the final pattern "-ем" or a prefix "под-" for example. We can assume that once we have a POS corpus of Surzhyk we can develop the rules in accordance with the part-of-speech characteristics and these would allow to reduce the number of the unappropriate results, or even minimize the effectiveness of the Surzhyk identification rules when applied on Russian corpus. But we have to consider that Surzhyk is mixing of Russian and Ukrainian standards and for this reason presenting elements of both languages is an important characteristic of its nature, that can explain why some rules on its identification are also efficient when applied on standard Russian texts.

## 7. Conclusions

The purpose of this research was to demonstrate whether it is possible to identify the elements of a hybrid Ukrainian-Russian language Surzhyk automatically. We focused our study on the particular group of Surzhyk-characterising elements we discovered in the analysed Surzhyk samples: the first person plural verbs of Surzhyk in Present tense and the Surzhyk verbs with a prefix "под-". The research was conducted with an example-based method. Through the creation of the written corpus we studied the Surzhyk samples and defined the rules for pattern identification of hybrid verbs. The rules were then implemented and tested using the R language.

We studied the theories on language contact and its consequences, in particular the case of Surzhyk. The acquaintance with the basis of Natural Language Processing and the study of Surzhyk samples allowed us to design effective rules and to automatically query the Surzhyk corpus. By adopting an example-based method we created rules on the automatic identification of the first person plural verbs in Present tense, and of verbs with a prefix "под-". We prepared a list of false positive results; it was expected that during the testing phase, independently of their morphological affiliation, all the words presenting a pattern corresponding to our search could emerge. For the automatic identification of verbs with a prefix "под-" we have 1 general rule, while for the first person plural verbs identification we created two groups of rules: general and specific. The adoption of the specific rules led to the successful identification of the aimed elements and reduced the number of false positive results we had with general rules. Moreover, the study of the endings of the first person plural of Surzhyk has demonstrated that there is an internal coherence in the Surzhyk verb phrase. Since this internal linguistic consistency is limited to our corpus, we propose to verify the hypothesis of internal coherence in Surzhyk verb phrase.

The identified elements concerning verbs with the ending of the first person plural were 11, while these regarding the verbs with a "под-" prefix were 4. The total amount of non-standard terms in the analysed texts were 1408. We should point out that not every entity that is part of 1408 may be considered Surzhyk, but surely more than half of these terms appertain to Surzhyk. Further studies in this direction should develop towards Part of Speech Tagging (POST). During the process of analysis it was realised that different types of rules could work better when applied on the already classified entities. Studies on POST are common for standard languages, but when in regard to non-standard languages they are not developed enough. At this point we cannot provide an annotated corpus to the non-standard language in question but we provide seven simplified terminological tables of Surzhyk with a total amount of 1408 terms, a model of the deeper description of the hybrid language terms standard terminological table of Surzhyk with 5 entities, 16 specific rules on the automatic identification of the final pattern of the first person plural ending in Present tense and one general rule on the identification of Surzyk verbs with a "под-" prefix implemented in R.

## 9. Bibliographical References